*Article*

# Virtual and Remote Robotic Laboratory Using EJS, MATLAB and LabVIEW

**Dictino Chaos** [1,*], **Jesús Chacón** [1], **Jose Antonio Lopez-Orozco** [2] **and Sebastián Dormido** [1]

[1] Department of Computer Science and Automatic Control, UNED, Juan del Rosal 16, Madrid 28040, Spain; E-Mails: jchacon@bec.uned.es (J.C.); sdormido@dia.uned.es (S.D.)

[2] Department of Computers Architecture and Automatic Control, Complutense University, Ciudad Universitaria, Madrid 28040, Spain; E-Mail: jalo@dacya.ucm.es

\* Author to whom correspondence should be addressed; E-Mail: dchaos@dia.uned.es; Tel.: +34-913-987-157.

**Abstract:** This paper describes the design and implementation of a virtual and remote laboratory based on Easy Java Simulations (EJS) and LabVIEW. The main application of this laboratory is to improve the study of sensors in Mobile Robotics, dealing with the problems that arise on the real world experiments. This laboratory allows the user to work from their homes, tele-operating a real robot that takes measurements from its sensors in order to obtain a map of its environment. In addition, the application allows interacting with a robot simulation (virtual laboratory) or with a real robot (remote laboratory), with the same simple and intuitive graphical user interface in EJS. Thus, students can develop signal processing and control algorithms for the robot in simulation and then deploy them on the real robot for testing purposes. Practical examples of application of the laboratory on the inter-University Master of Systems Engineering and Automatic Control are presented.

## 1. Introduction

In Robotics education, especially in e-learning environments in higher education, it is very important to use laboratories that allow to practice the acquired knowledge and to observe the real errors or problems that can appear in a real system but do not exist in simulated or virtual laboratories. An interesting approach are the web-based laboratories [1,2]. In this methodology, remote laboratories are e-learning resources that enhance the accessibility of experimental setups, providing a distance teaching framework that meets the users' hands-on learning needs.

Nowadays, remote laboratories can be integrated into a Learning Management System (LMS) or a Content Management System (CMS), where they can be seen as additional resources available for the users. These laboratories have allowed to make new practices in the context of the course. To ensure an exclusive access control of one user at any time, it is necessary to use a booking system integrated within the learning environment system. Examples of robotic remote laboratories with a booking system are [3–5].

This paper shows a virtual and remote laboratory designed for the students of the Autonomous Robots subject in the Master of Systems Engineering and Automatic Control of UNED and Complutense University. This Master is based on the distance education paradigm [6], where all subjects use e-learning methodology. This means that students from different countries (mainly from Spain and Latin America) study from their own homes.

E-learning methodology is especially difficult to apply to practical subjects like Autonomous Robots. There are some problems that arise on the use of remote laboratories. First, if the interface of the laboratory has not been designed carefully and the programming language is not well known by students, the implementation of the practice becomes complicated. Second, only a few robots are available because the equipment necessary to construct it is generally expensive [7]. Thirdly, a schedule must be taken into account because in robotic laboratories there are more students than physical resources.

In order to solve the first problem MATLAB and EJS are used. MATLAB is a well-known tool widely used in remote laboratories [8,9]. On the other hand, EJS [10] is an authoring tool that can be used for developing virtual and remote laboratories [11,12]. The cost problem is overcome using a LEGO robot. Examples of similar laboratories include: [13], where a LEGO robot is controlled with MATLAB; [14], where a Surveyor SRV-1 robot is remotely controlled using MATLAB and EJS; [15], where a set of mobile robots and wireless sensors are controlled using Tiny-OS; and [16], where a set of remote laboratories are available for Control Engineering subjects. Finally, the scheduling of the control access is achieved since these practices are included in a repository named UNEDLabs [17] where students can find laboratories of different subjects. This repository is a Moodle LMS, which offers a booking system and a collaborative support.

Besides these problems, the objectives of this project are threefold: (1) to solve the problem of the distance learning by means of developing a remote laboratory that allows the students to access to a real robot and carry out some experiments from their own homes using an Internet connection; (2) to use a Java interface like EJS that is known to all students; and (3) to create a similar interface for virtual and remote laboratories.

A virtual and remote laboratory has been implemented in order to simulate the robot before doing the real experiments. Both laboratories have the same interface integrated in Moodle course. The use of the virtual laboratory in combination with the remote laboratory with the same interface has many advantages. First, it makes an efficient use of the resources because the student can learn how the laboratory works prior to use the real robot. Therefore they can take the most of the time they have been assigned with the real robot because they already know how to operate it. This is very useful in the first phase where learning is dominated by a trial and error process and also in the programming phase to make a first test of the user code before evaluating it in the real setup.

The second advantage is that it makes the timetable more flexible because the simulations are available all the time so students can adapt their schedule to perform the experiments. It must be noticed that during the real experiments it is necessary to have a tutor that supervises students' work in order to evaluate the practice and solve technical problems that may arise in the experimental session.

The third advantage is that students have a lot of freedom to perform any experiment they want with the virtual lab without any danger for the real system. Thus they can get confidence in their work before testing on the real robot [18]. This increases the creativity of students during the design stage because in general they are more conservative when they feel that the laboratory can be damaged.

Finally it is important to remark that the virtual lab cannot substitute the real laboratory since it is almost impossible to make a perfect model of the real system. Despite of that, the virtual laboratory performs a simulation of a simplified model of the robot that students can interact with in the same manner that with the real robot.

The rest of the paper is organized as follows: Section 2 explains the architecture of the virtual and remote laboratory (server and client sides) in order to use the laboratory like a stand-alone application or an applet embedded in Moodle. Section 3 shows a practice performed by a student in the UNEDLabs LMS: the results in the virtual/remote laboratory using the same EJS interface. Section 4 contains our conclusions regarding this work.

## 2. Virtual and Remote Laboratory

The virtual and remote laboratory can be decomposed in two separated parts as depicted in Figure 1. The first part is in the client side that is developed in Java using the authoring tool EJS [10]. This is the part that is employed by students and defines their interactions with the laboratory. Students can access the laboratory using a stand-alone application or an applet embedded in Moodle.

**Figure 1.** Laboratory Architecture.

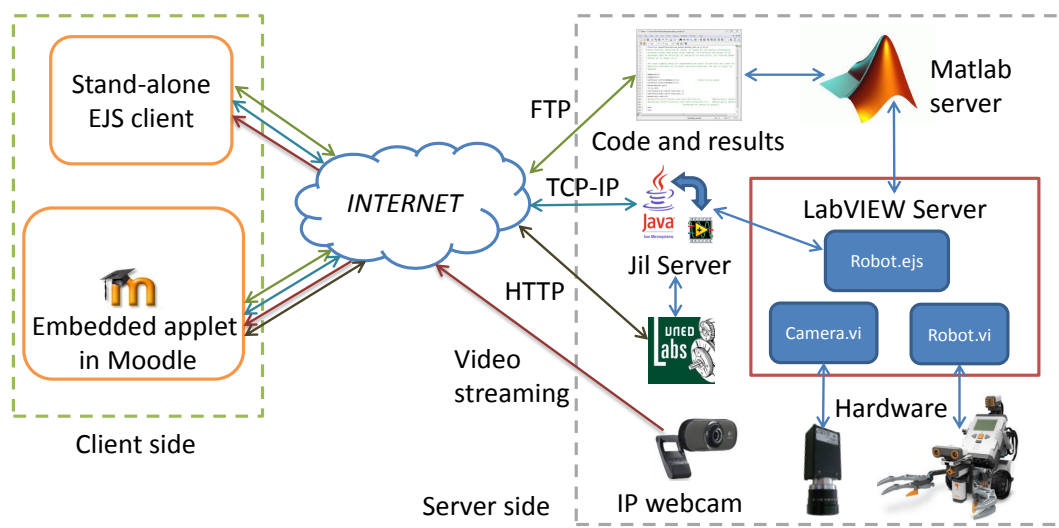

In the first case the interface is a Java executable that directly exchanges information with the laboratory using three protocols: FTP to load code to the server and save results, TCP-IP to communicate with Jil Server, and video streaming from the Webcam (available only on the remote laboratory).

In the second case the mechanism is the interaction with an applet that is placed on a Moodle course, so in addition to the protocols described above, the browser will exchange information with the server UNEDLabs that implements the Moodle courses.

The second part is the server side and it is composed by the hardware and software elements. Hardware components are the robot, the cameras for the robot positioning, the IP Webcam and the computers that host the server. On the other hand, software pieces are: the MATLAB server that executes the student code, hardware controllers for each element of the laboratory, Jil Server [19] that communicates LabVIEW with Java clients, UNEDLabs server that hosts the course, and the booking system.

In parallel with the main task, some services are running. In particular, a FTP server is used to exchange code and data with the robot and an IP Webcam is used to provide visual feedback.

In the following, each element will be described in detail.

## 2.1. Server

This part of the application implements both the real laboratory and the simulated one. The server is composed of four physical elements, the LEGO NTX robot that is tele-operated using Bluetooth, the high resolution camera that computes the position of the robot over the floor, the Webcam that gives to the student the visual feedback of the robot operation, and the computer that integrates the information from the different sensors, executes the user code and sends the commands to the robot. The experimental setup is depicted in Figure 2.

**Figure 2.** Server side of the laboratory.

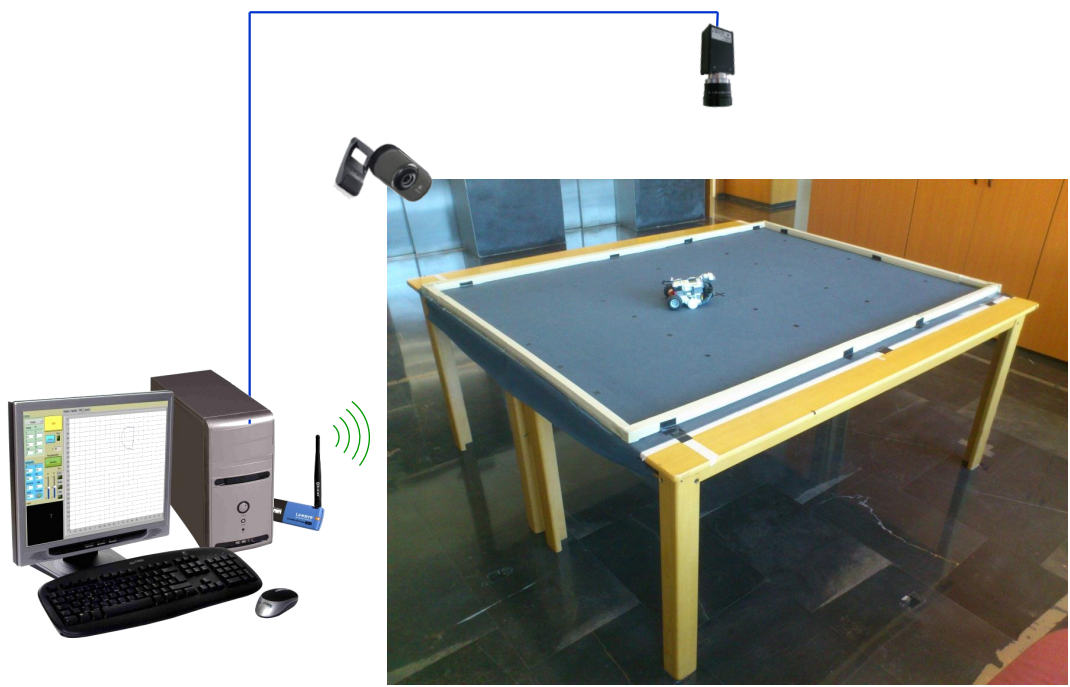

The server is implemented in the LabVIEW programming language. This graphical language was selected because it allows direct connection with the LEGO NXT robot and gives advanced image processing capabilities that are used to compute the robot position.

The server consists of three parts: `Camera.vi`, `Robot.vi` and `RobotEJS.vi`. These parts communicate with each other using the global variables `Robot_inputs`, `Camera_Positions` and `Robot_outputs`.

2.1.1. Lego Robot

Lego Mindstorms NXT is a programmable robotics kit. The main component in the kit is the NXT intelligent brick, which is based on a 32-bit ARM7 processor. It can control up to three motors and receive inputs from up to four sensors, and it has support for Bluetooth wireless communication.

LEGO Mindstorms NXT provides a range of sensors [20] that cover the commonest applications in robotics, including (but not limited to)

- A *Touch Sensor*, which can be used to detect physical contact of the robot with other objects.
- A *Light Sensor* can detect light intensity in a room and measure the light intensity of colored surfaces.
- A *Sound Sensor* can detect sound pressure, giving a measure either in decibels (dB) or adjusted decibel (dBA).
- An *Ultrasonic Sensor* to detect obstacles in a range from 0 to 255 cm.
- A *Compass Sensor* is able to calculate the robot direction by measuring the earth's magnetic field.
- A *Color Sensor* that can detect 6 different colors, read the light intensity in a room and measure the light intensity of colored surfaces.
- An *Accelerometer Sensor* can measure both the orientation and acceleration of the robot in three axes.

In addition, there are other vendors that provide advanced sensors such as a gyroscope and accelerometer, thermal infrared sensors or bending motion sensors [21].

The intelligent brick can be programmed with NXT-G, which is a graphical programming environment bundled with the NXT, but also admits a wide variety of languages, such as LabVIEW, C or Python.

Though the structure of a robot built with LEGO Mindstorm NXT can be adapted as desired to fulfill the requirements of the user application, a wide variety of ready-to-use designs can be found on the Internet. In addition, the LEGO Mindstorm NXT capabilities can also be easily extended by integrating with other hardware platforms (Arduino, to cite one example).

The main advantage of this robot is that it is very flexible and it can be reconfigured easily to use different kinds of sensors. In the following, we use one possible configuration of the system, built by following the basic structure described in the NXT User Guide (8547), with some minor modifications like adding three lights in order to be tracked by the artificial vision subsystem. It is composed of a chassis built with LEGO construction blocks, which carries the NXT intelligent brick and an ultrasonic sensor to detect obstacles, and two servo motors connected to wheels, which allow the robot to translate and/or rotate in the floor surface.

2.1.2. Robot Interface

The NXT robot is connected to LabVIEW using the *NXT LabVIEW Library* that allows to access the internal state of the robot and to execute commands using Bluetooth.

The module `Robot.vi` shown in Figure 3 controls the robot as follows. The outer loop tries to connect continuously to the robot and, once the connection is established, the inner loop is executed. This loop first turns on the lights of the robot, enabling the vision system to detect it. Then the global variable `Control_input` is read, obtaining the value of the control signal U. This signal is normalized to the maximum speed allowed on the experiments (that is limited to a 30% of the maximum motor speed for safety reasons) and then it is sent to the motors. Once the control signals are sent to the vehicle, the data of the sensors and the state of the battery are read from the robot. In this example, an ultrasound range sensor is used to measure the distance to the obstacle. Then the distance and the state of the battery are stored on the fields `Distance` and `Battery` of the global variable `robots_outputs`.

**Figure 3.** Robot.vi.

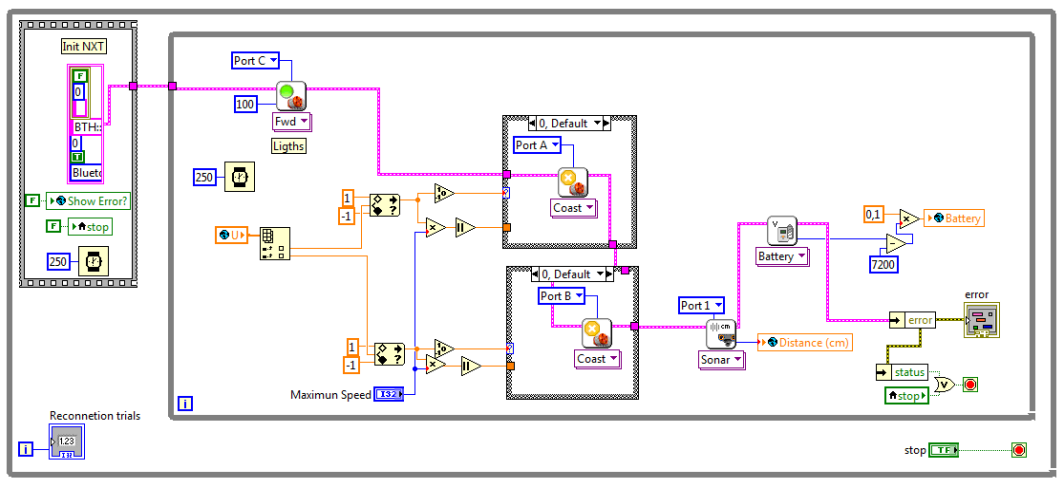

2.1.3. Artificial Vision Subsystem

In order to track the position of the robot, a high resolution CCD camera equipped with wide angle optics is placed 3 m over the platform where the robot moves. This camera can cover a region of approximately $3 \times 4$ m$^2$. The camera is attached to the server computer through a frame grabber that allows a frame rate of 17 fps.

The robot is equipped with three high bright LEDs that form an isosceles triangle, thus the module `Vision.vi` can compute its position as shown in Figure 4.

First, a video frame is captured by the `Vision Acquisition` Block. Then the image is processed by the block `Points.vi` and the centroid of each individual LED is computed. The next block compensates for the barrel distortion introduced by the lenses, obtaining the undistorted position of the LEDs on camera frame.

Once the corrected positions of the points are computed, the next block solves the perspective transformation from camera frame to floor coordinates, which finally yields the real positions of the lights on the surface where the robot moves. Since the positions of the LEDs are known, it is straightforward to compute the center of mass of the vehicle and its orientation from the positions of the triangle vertex, and then estimate the velocities by a finite difference.

Finally the positions, orientation and velocities are stored on the global variable `Camera_Positions` as well as the time and the error handler.

**Figure 4.** Camera.vi.

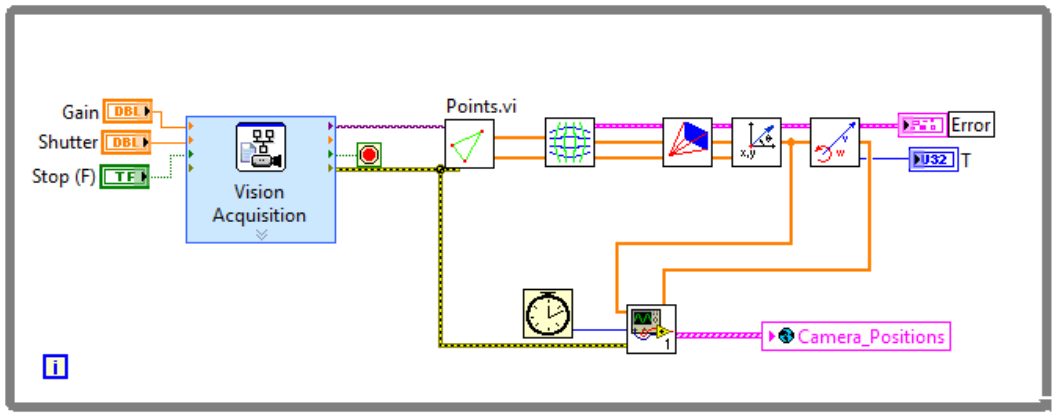

### 2.1.4. EJS Interface

The EJS interface is the core of the system. It integrates the measurements from the vision subsystems and the robot measurements, invokes the user code that is executed in a MATLAB server, and sends the resulting control inputs to the robot. The interface is implemented in `robotEJS`, shown in Figure 5, and can work in two modes of operation, the simulated one and the real one.

**Figure 5.** RobotEJS.vi.

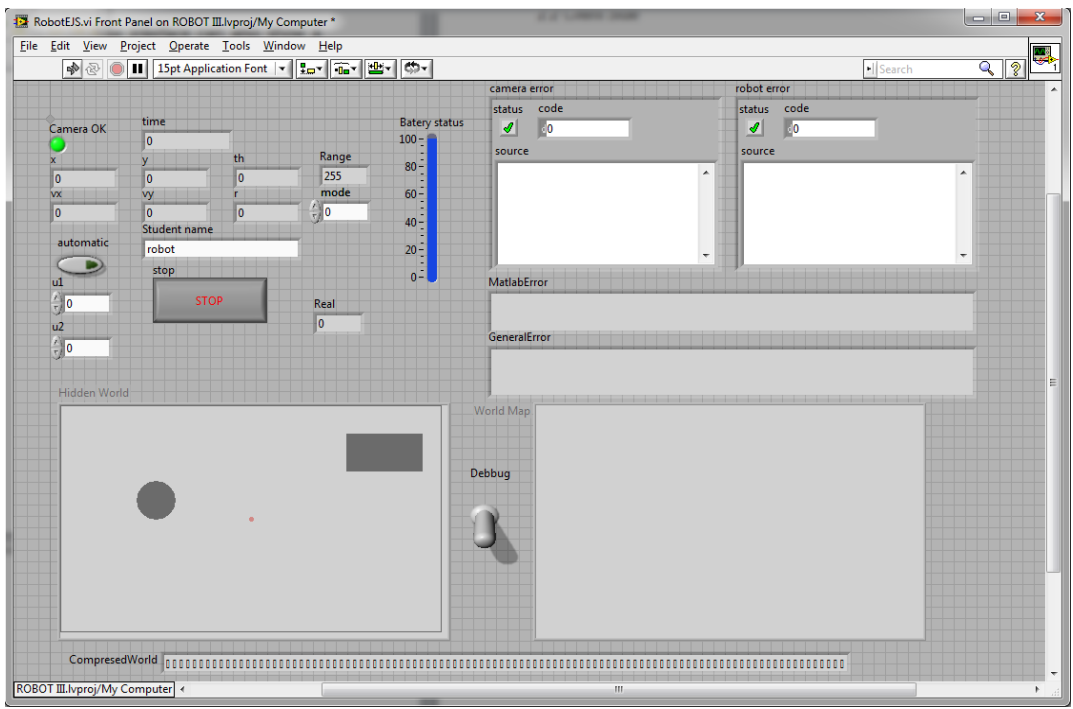

When the server is initiated in real mode, it first starts the MATLAB server. Then, it checks the student name and changes the path where the student code resides. This directory contains at least two scripts (`init.m` and `close.m`) and two functions (`ComputeWorld.m` and `ComputeControl.m`) written in M language.

The `init.m` script creates the `world` matrix, which is a 300 × 400 matrix on the MATLAB workspace that represents the environment of the robot. The value of each pixel represents the occupancy of each cell. Depending on the mapping algorithm, the occupancy of a cell could be deterministic (occupied/void) or probabilistic, in the later case the value of the pixel represents the probability of been occupied based on the measurements of the systems (*i.e.*, Bayesian mapping). Each 200 milliseconds, the server executes the function `ComputeWorld(world,x,y,th,d)` that updates the matrix of occupancy, `world`, according to the position of the robots `x` and `y`, the orientation `th`, and the distance to the obstacle `d`.

Next, if the control is in automatic mode, the server invokes the function `control.m` (`[u1,u2]=control(world,x,y,th,vx,vy,w,d,t)`) where `vx,vy` and `w` are the velocities and angular velocity estimated by the vision subsystem, and `t` is the time. This function computes the control action `u1` and `u2` based on the state of the vehicle, the time and the representation of the robot environment.

In case that the robot is in manual mode, the server simply writes the values of the control action that are sent from the client to the control variable `U`. Thus the module `Robot.vi` will send these values to the robot.

When the server runs on simulation mode, everything works in the same manner except that `Robot.vi` and `Camera.vi` are not executed. Despite of that, the server calls a model of the system that computes the new state of the robot and the distance to the obstacles according to a hidden world model.

During the execution of the server, all the state variables are stored in a historic vector, *i.e.*, the position `x` is stored on a vector `x_hist`. When the server finishes its execution, it calls the `close.m` script that saves by default all the state variables and the resulting world representation on a MATLAB file `hist_DATE_TIME.mat`. The resulting data file is in .mat format, which can be opened by the students allowing offline processing.

The front panel of RobotEJS.vi is shown in Figure 5 and contains the state of the system, that is, the positions, the velocities, the state of the battery, the error and the `world` matrix. In addition, for debugging purposes, the interface can also show a graphical representation of the robot environment based on the robot position, and the `world` matrix. Furthermore, the hidden matrix that contains the virtual world used in simulation is also shown.

2.1.5. JiL Server

The TCP/IP communication between Java applets and LabVIEW can be implemented directly with the Java API and the LabVIEW communications library. Though it can be very effective, this approach is inconvenient in that the developed applications are frequently only known by the programmer, which degrades the maintenance and scalability of the system.

This mechanism is based on a middleware layer composed of a server, named JiL Server (Java Internet LabVIEW Server), which is located between the Java applets and the LabVIEW applications that perform the local real-time control in the laboratory.

JiL server acts as an interface between EJS and LabVIEW. It provides an easy way to send and receive the values of variables directly without entering into low level details. JiL provides a Java API to define in EJS variables associated to the controls and indicators of LabVIEW Virtual Instruments (VIs), and to control the load and the execution of these VIs.

The functions that are implemented in the LabVIEW application are divided into several areas, namely:

- *Data Acquisition*, the communication with the hardware of the robot, in order to read the data provided by the sensors of the robots and the video camera, and also to send commands to the actuators.
- *Simulation*, in this area the implementation of the simulation model used in the virtual laboratory is included.
- *Data Logging*, this subsystem contains the blocks that perform the writing of the data logged from the system to disk.

The reception of data is synchronous and managed by a continuous loop, where the view is updated with every new dataset sent periodically from the server. On the other hand, the commands are sent asynchronously, only when the student interaction with the view requires a communication with the server. This is done in this way to obtain a lower latency and therefore a better experience for the user. Latency depends on the network, the amount of information that must be sent and the time that JiL server needs to process the data. With this strategy, the time that JiL needs to execute commands (*i.e.*, open, run, close, stop a vi) is less than 1 ms, and the maximum amount of time that the JiL server needs to collect and send data is less than 20 ms.

In this application, the JiL Server checks user credentials against the booking system of UNEDLabs and, if they are correct, executes by demand the module `RobotEJS.vi` and synchronizes its variables with Java. Using JiL, LabVIEW variables can be accessed from the client application developed in EJS.

All the variables shown in Figure 5 are synchronized with the client with the exception of the indicator `Hidden World` that must be unknown for the student, `debug` that is used for debugging purposes, and the indicator `World Map` that shows the occupancy matrix `world`. In order to achieve a fluid communication between the client and the server, the `world` matrix is binarized and compressed. This compressed version of the `world` matrix is stored in the `CompresedWorld` and then exchanged with EJS.

2.1.6. Rest of Services

In addition to the server developed in LabVIEW that allows to control the robot of the laboratory, the PC runs a FTP service that is used by students to upload their code to the simulator. Each student has a username, a password and a work directory that is set up in the server during the connection.

The server verifies the user and password and it only allows the connection if the user is registered in the booking system for that time slot. The code that the student executes over the robot or the simulation resides in their own work directory and the results of the experiments and the simulation can be easily accessed via FTP from the application.

On the other hand, the visual feedback of the user of the laboratory is obtained using an IP Webcam whose operation is independent of the LabVIEW server. This is done in this way in order to separate the time critical part (implemented on the LabVIEW server) that must operate in real time from the visual feedback that is optional and can be disconnected in case of a poor Internet connection.

*2.2. Client*

The Graphical User Interface (GUI) has been developed by means of the software tool Easy Java Simulations (EJS). The interface brings to the students the possibility of interacting with the remote and/or virtual laboratories. EJS shows in a graphical and intuitive way the information of all the variables associated with the state of the robot and the measures obtained from its sensors, and it provides a means of sending commands to control the execution of the experiments required in the exercise. In the following section, the software tool used to develop the interface, namely EJS, is briefly reviewed. Afterwards, the architecture and design of the GUI is explained in detail.

2.2.1. EJS

Easy Java Simulations (EJS) [10] is an open source (and completely free) software tool designed to create simulations in Java with high-level graphical capabilities and with an increased degree of interactivity. The tool provides its own mechanism for describing models of scientific and control engineering phenomena, and, as such, it can be used to create virtual laboratories in a very simple and efficient way. The definition of the model can be done either in an *ODE page*, introducing the differential equations with the dynamics of the system, or in an *evolution page* (Figure 6(a)), where the introduced code is executed periodically, reading and updating the variables of the model (Figure 6(b)).

**Figure 6.** Definition of the model in EJS. (**a**) Subpanel Evolution of EJS; (**b**) Subpanel Variables of EJS.

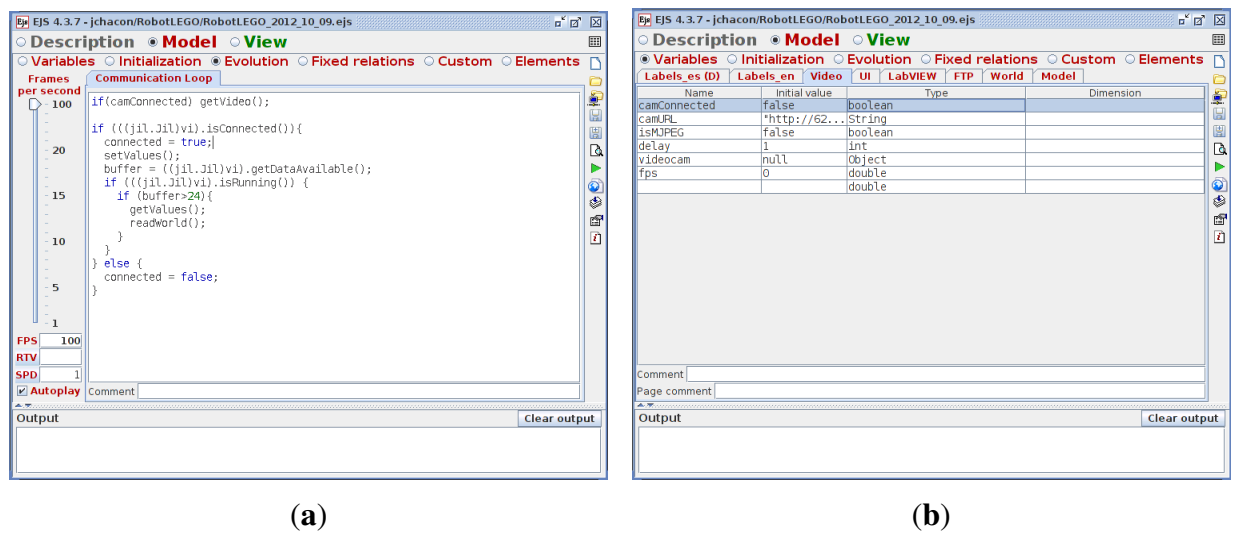

(**a**) (**b**)

The design of a simulation in EJS is based on the MVC (*Model-View-Controller*) paradigm. The main idea of this approach is the division of the system in three elements conceptually differentiated: the *model*, which is an internal representation of the problem; the *view*, whose purpose is to present the state of the model in a proper way, and the *controller*, which responds to the user interactions and translates them into requests for the model. The tool has been conceived for science students and teachers, that is, for people who are more interested in the content of the simulation, the simulated phenomenon itself, and much less in the technical aspects needed to build the simulation.

2.2.2. User Interface

The information is presented to the students in two complementary ways: all numeric values are shown in a textual way so that they can know in every time instant the exact state of the robots; in addition, there is a graphical representation of the robot's world, which is dynamically updated with the data received, as shown in Figure 7. Finally, a view of the real environment of the robot obtained from a Webcam is at the student's disposal.

**Figure 7.** The GUI of the remote and virtual laboratory in EJS.

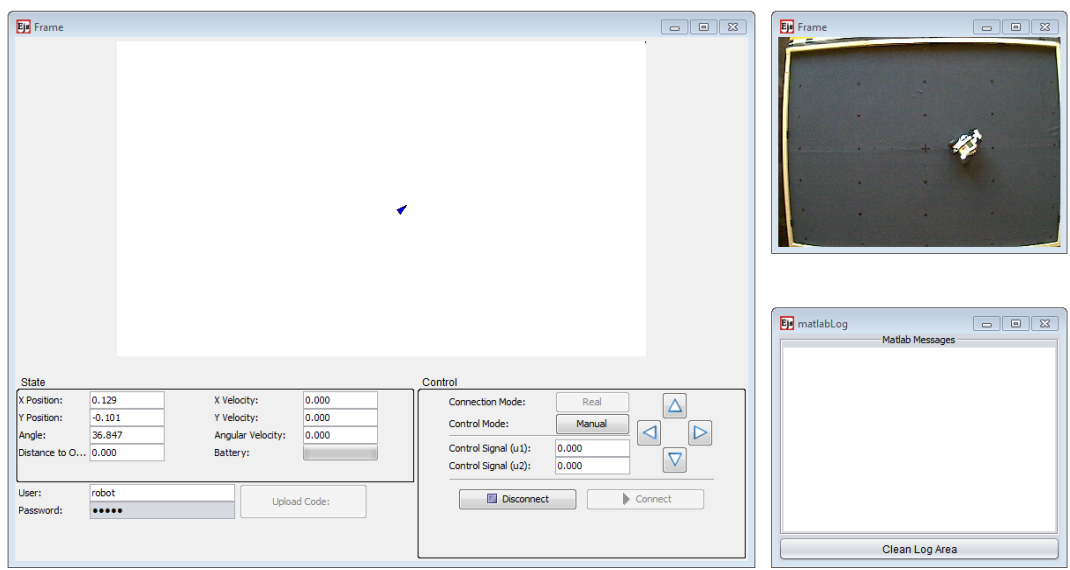

The application allows for two modes of execution. The practice session can be done in the remote laboratory, where the LEGO robot is directly controlled and the results are obtained from real measures of the sensors. Alternatively, the practice session can also be done in the virtual laboratory, where there is a simulation server that generates simulated measures in response to the student interaction and/or the uploaded code.

In the control zone, the user can configure the execution mode by means of a button that switches between real mode and simulation mode (virtual/remote lab) as well as to connect to or disconnect from the server. Once connected to the server, it is possible to take manual control of the robot, which allows to send the movement commands directly to it by using the arrow buttons, or to execute it in the automatic mode, where the behavior of the robot is determined by the controller code written in MATLAB, which must have been previously uploaded with the button provided for that purpose.

Finally, it is worth to mention again that access control is required, *i.e.*, the student must log in with a valid username and password that is checked by the JiL server and allows them to upload the code files to their assigned space on the server, and to download the data files with the results of their working sessions.

## 3. Experiments with the Laboratory

The virtual and remote laboratory can be used on two modes as depicted in Figure 1: students can access to the laboratory using a stand-alone application or an applet embedded in Moodle. In the first case, the interface is Java executable, an EJS application, which directly exchanges information with the laboratory. In the second case, students interact with an applet that is placed on a Moodle course, so in addition to using the communication protocols like the stand-alone application, it will exchange information with the server UNEDLabs. Hereinafter, an example of the laboratory in the Moodle platform is shown.

### 3.1. Integration with Moodle

The client consists of an EJS jar file that contains all the resources needed to execute the remote laboratory and the virtual laboratory. They have the same interface, so students can train how to use the LEGO robot with the virtual environment before they connect to the real robot.

The UNEDLabs [17] Moodle LMS allows to include all the resources (scripts, user guides, theoretical documents, ...) along with the jar file of the laboratory. Beside, this LMS provides a booking system in order to control the access of students.

The virtual and remote laboratory have been tested with the students of the subject Autonomous Robots that attend to 25–30 students at year. The use of the laboratory consists in a practice included on the subject Computer Science and Robotics of the Master of Systems Engineering and Automatic Control of UNED and Complutense University.

The main objective of the practice is that students solve the problem of robotic mapping, which is to create a map of the environment of the robot based on the measurements of the range sensors and the position of the vehicle. For that purpose students must implement `ComputeWorld.m` in MATLAB language, which has been selected because it is a standard for signal processing and control and also because students use this language in other practices of the Master.

The work that students have to do consists of two phases, the virtual experimentation and the real tests. The structure of the practice established that the student should have successful results on simulation prior to have access to the real system. In Figure 8, the Autonomous Robots course into the UNEDLabs LMS is shown. Two phases can be seen: the phase I is the virtual laboratory and the phase II is the remote laboratory. In phase I, the *practice1* link is a script where the homework of students is described. These jobs must be completed using the virtual laboratory before students can move to the remote laboratory. In phase II, the *practice2* link shows what students should do with the real robot. There is only one laboratory link available because the same applet is used for the two phases.

**Figure 8.** UNEDLabs Website and Autonomous Robots course.

## 3.2. Virtual Laboratory

During the first stage of the practice students must deal with a simulation of the robot using the interface described in Section 2.2. The simulation is implemented on the LabVIEW server. This implies that students need an Internet connection to do the experiments.

During this phase students control the robot manually and can implement any algorithm in MATLAB to create an automatic mapping. The only constraint imposed to students is that the code that updates the map must be executed in 200 ms on the server.

In this practice students may discover a hidden map of the environment. In order to enhance the learning process, an example implementation of the mapping algorithm is accessible to be executed. To use this example students must login with the *example* user. Nevertheless, it must be noticed that the example user code is not accessible via FTP so students cannot see or modify it. In addition, the hidden environment is different for each user so the results of the example user are not the solution of the problem that students must solve.

Once this part of the practice is completed, the results and the code are sent to the professor to be evaluated. A successful example of a mapping algorithm developed by a student, which has been tested in simulation, is shown in Figure 9. As the figure shows, the student code is able to reconstruct the map of a hidden world composed by a circular and rectangular shape with great accuracy.

**Figure 9.** Example of a successful execution of a student code.

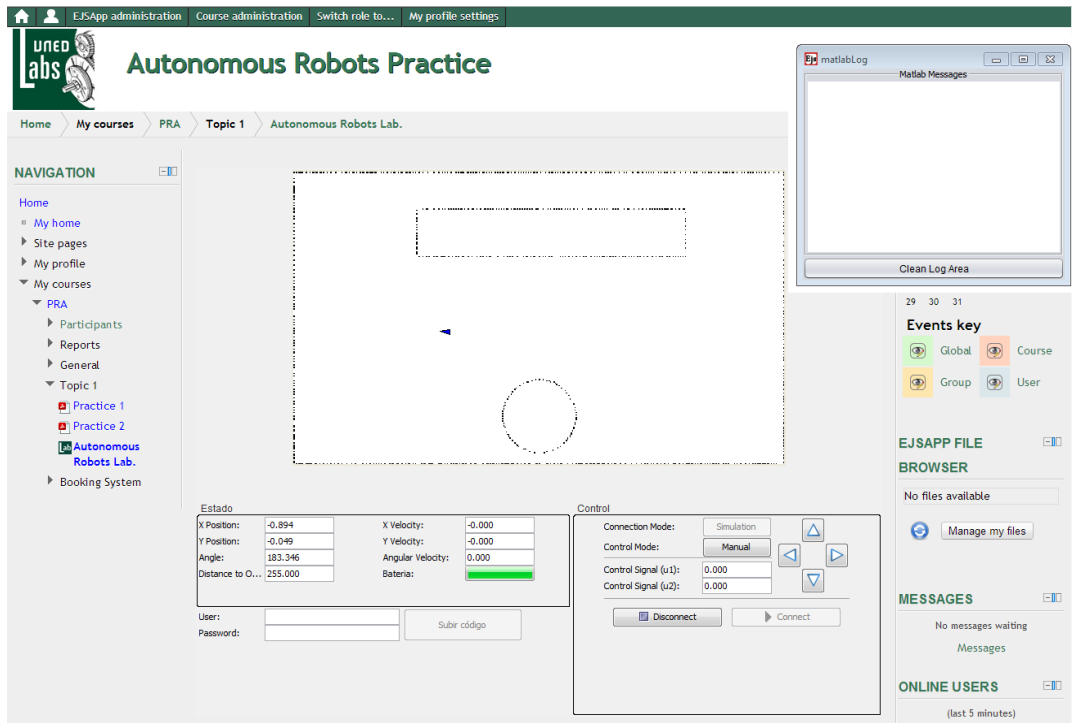

## 3.3. Remote Laboratory

When the simulation results are correct, the professor assigns the student a time slot in the real laboratory to test their code on the real vehicle.

During the operation of the real vehicle, some assistance is needed because it is possible to have some problems that cannot be solved remotely by the student, such as loss of communication with the robot, mechanical failures, collisions, *etc*.

From the user interface point of view, operation with the real laboratory is almost the same as with the virtual laboratory. The main difference between the real laboratory and the remote one is that in the second case the user has visual feedback of the laboratory, as Figure 10 shows.

Nevertheless, the experiments with the real plant go beyond the straightforward test of the algorithms developed in simulation. This implies that, during the test, students need to make changes on their code. The main differences between the virtual and remote laboratory are the following:

- The model does not have the same complexity as the real robot. The motor suffers from nonlinear effects like saturation on acceleration and dead zones, and in some operations wheel slippage can be observed.
- Sensors are not perfect but are affected by noise and outliers on the measurements.
- Delays that does not appear in the simulations exist in the measurements of the different sensors. In our setup, the camera compute positions each 59 ms and the Bluetooth communication with the robot introduces a delay of 250 ms on the range measurements of the ultrasonic sensor.
- The robot becomes harder to control both in manual mode and in automatic mode due to the delay of 250 ms introduced on the control action by the Bluetooth connection.

- Sometimes there are temporal failures on the communication and some measurements from the ultrasonic sensor and commands sent to the robot are lost. This can lead to mechanical problems of the vehicle, like collisions with obstacles.

**Figure 10.** Example of a real experiment performed by a student.

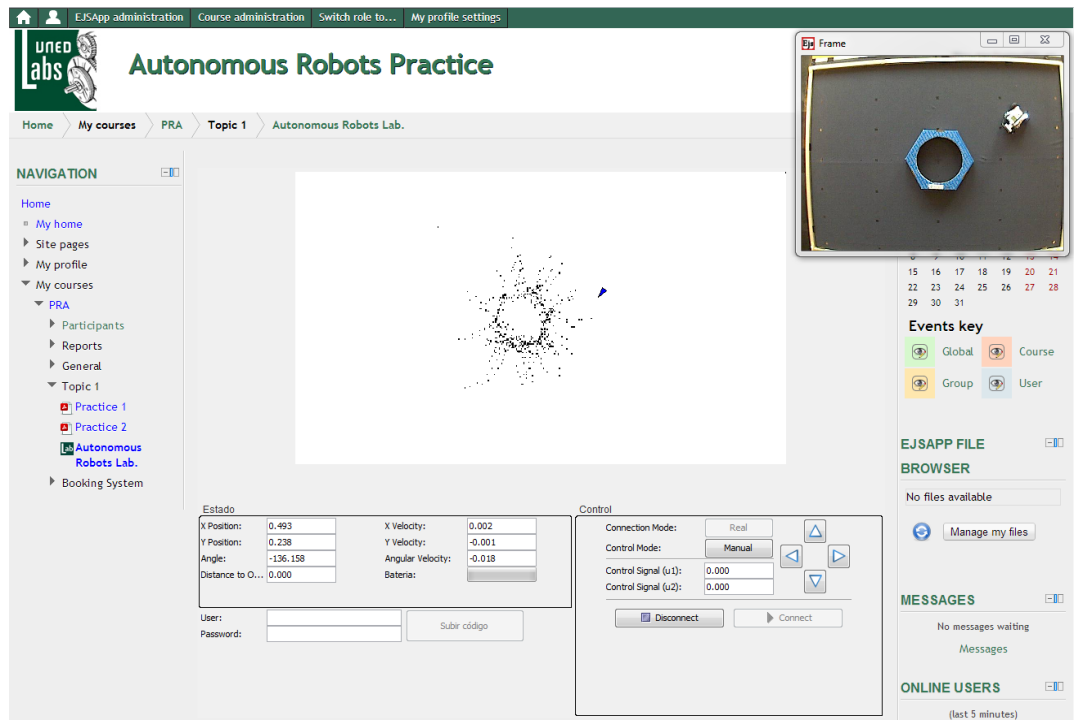

This implies that the real experiments are a rich source of knowledge to enhance the learning process of students that must deal with practical problems usually neglected in theory but arise in the sensors of real robots.

*3.4. Other Experiments*

In the Sections 3.2 and 3.3, a practice carried out by students is shown. In this case, students obtain a map from noisy and uncertain sensor measurement data. They use an ultrasonic sensor in order to build the map of its environment with an Occupancy Grid Mapping method. This method assumes that the robot position is known, which is provided by the artificial vision subsystem described in Section 2.1.3.

Due to the flexibility of the robot, other configurations of the sensors are possible, yielding interesting practices that make use of other types of sensors, namely:

- Multi-sensor fusion. In this practice, ultrasonic and infrared sensors are used to get a better evidence of the presence of an obstacle in the environment. An occupancy grid is used to compute the belief of the detected objects. This exercise shows that the map obtained by fusing data of several sensors is better than the one built using one sensor.
- Inertial navigation and Kalman filtering. In this case, the initial position of robot is known and the students, by means of dead reckoning, use the motion sensors (accelerometers) and rotation sensors (gyroscopes) to calculate the position, orientation, and velocity of the moving robot without the need for external references. However, small errors in the measurement of acceleration or orientation are accumulated roughly proportionally to the time. Therefore the position must be periodically corrected by input from some other type of measure. In our case, the artificial vision subsystem is used like a GPS in order to reduce the position error. So, a Kalman filter provides a theoretical framework for combining information from both sensor systems to improve the estimate of the robot position.

- Odometric navigation and Kalman filtering. This is a variant of the last practice in which the encoders of the robot are used to know its position during the navigation instead of the inertial sensors.

These exercises, and others as SLAM (Simultaneous Localization and Mapping), are offered to students and they can select the practice they want to implement.

*3.5. Offline Work and Evaluation*

After the experiments with the real system, students must do a report with the main results obtained in simulation and with the experiments. During this stage, students must think about the problems of the real systems and propose new algorithms to deal with this issue.

Students can download the history of the experiments (using FTP or the Moodle workspace). They are encouraged to reconstruct by themselves and solve the problems that appear on the real system.

The final report must show the simulation results, the experimental ones and also the proposed enhancements of the mapping method. This must be sent to the professor to be evaluated. During the evaluation, the professor pays special attention to the original solutions proposed by students.

The main reason is that if the student is capable of enhancing the original algorithms, this implies that the learning process was successful and the student got a deep understanding of the underlying problems of sensors presented on mobile robotics in real applications.

## 4. Conclusions and Future Work

The virtual and remote laboratories have been operating in the Master of Systems Engineering and Automatic Control of UNED and Complutense University for two years. In these years, it has shown to be a very useful tool for teaching the role that sensors play on robotics. Student satisfaction survey shows that they agree or completely agree with the fact that the laboratory is necessary for a complete understanding of the robotic sensors.

Using the virtual laboratory like the remote one has been a success: students can get confidence in their work before testing in the real robot and they can make a first debug of the code before testing it in the real robot; 74% of the students feel more comfortable if they can use the virtual laboratory before connecting with the real robot.

Students can access the laboratory using a stand-alone application or an applet embedded in Moodle. In the first case, the interface is a Java application that directly exchanges information with the server of the laboratory and the results are directly saved by students and are sent to the professor. In the second case, the student interacts with an applet that is placed on a Moodle course and the results are saved and located in his workspace of Moodle where the professor can get them.

The Laboratory setup is very flexible and this paper only explains in detail one example of use of this robot. There exist many other configurations that can be used by the students employing different configurations of sensors, *i.e.*, computing the position of the robot with odometer and comparing with artificial vision, or implementing an inertial navigation system based on a compass sensor and accelerometers.

In addition, new control problems will be included in the laboratory, like position control and path planning based on the measurements obtained with the sensors.

## Acknowledgments

The authors want to thank the Ministry of Economy and Competitiveness of Spain for the support received from grants DPI2009-14552-C02 and DPI2007-61068.